\documentclass[final]{article}

\usepackage{neurips_2025}

\usepackage[utf8]{inputenc} 
\usepackage[T1]{fontenc}    
\usepackage{hyperref}       
\usepackage{url}            
\usepackage{booktabs}       
\usepackage{amsfonts}       
\usepackage{nicefrac}       
\usepackage{microtype}      
\usepackage{xcolor}         
\usepackage{graphicx}
\usepackage{amsmath}
\usepackage{natbib}
\usepackage{float}
\newtheorem{theorem}{Theorem}[section]

\newtheorem{lemma}[theorem]{Lemma}
\newtheorem{definition}{Definition}[section]
\newtheorem{assumption}{Assumption}[section]
 \usepackage{amssymb}
 \usepackage{caption}
\title{Permutation Randomization on Nonsmooth Nonconvex Optimization: A Theoretical and Experimental Study}

\author{%
  Wei Zhang\thanks{The corresponding author.}\\
  School of Computer and Cyber Sciences \\
  Augusta University, Augusta, GA, USA  \\
  \texttt{wzhang2@augusta.edu} \\
  \And
  Arif Hassan Zidan \\
  School of Computer and Cyber Sciences \\
  Augusta University, Augusta, GA, USA \\
  \texttt{azidan@augusta.edu} \\
  \And
   Afrar Jahin \\
   School of Computer and Cyber Sciences\\
   Augusta University, Augusta, GA, USA\\
  \texttt{ajahin@augusta.edu} \\
  \And
  Yu Bao \\
  Department of Graduate Psychology \\
  James Madison University, Harrisonburg, VA, USA \\
  \texttt{bao2yx@jmu.edu} \\
  \And
  Tianming Liu \\
  School of Computing \\
  University of Georgia, Athens, GA, USA \\
  \texttt{tliu@uga.edu} \\
}

\begin{document}

\maketitle

\begin{abstract}
While gradient-based optimizers that incorporate randomization often showcase superior performance on complex optimization, the theoretical foundations underlying this superiority remain insufficiently understood. A particularly pressing question has emerged: \textit{What is the role of randomization in dimension-free nonsmooth nonconvex optimization}? To address this gap, we investigate the theoretical and empirical impact of permutation randomization within gradient-based optimization frameworks, using it as a representative case to explore broader implications. From a theoretical perspective, our analyses reveal that permutation randomization disrupts the shrinkage behavior of gradient-based optimizers, facilitating continuous convergence toward the global optimum given a sufficiently large number of iterations. Additionally, we prove that permutation randomization can preserve the convergence rate of the underlying optimizer. On the empirical side, we conduct extensive numerical experiments comparing permutation-randomized optimizer against three baseline methods. These experiments span tasks such as training deep neural networks with stacked architectures and optimizing noisy objective functions. The results not only corroborate our theoretical insights but also highlight the practical benefits of permutation randomization. In summary, this work delivers both rigorous theoretical justification and compelling empirical evidence for the effectiveness of permutation randomization. Our findings and evidence lay a foundation for extending analytics to encompass a wide array of randomization.

\end{abstract}

\section{Introduction}

Optimization problems in machine learning are typically addressed using numerical optimizers that rely on either full gradients, computed over the entire dataset, or stochastic gradients, calculated from individual data points or mini-batches. Full-gradient methods deliver theoretical guarantees of eventual convergence, while stochastic gradient methods provide advantages in terms of convergence speed and computational efficiency~\citep{Hazan2007logarithmic, Nemirovski2009robuststochastic, jordan2023deterministic}. In parallel, a growing body of research has explored hybrid optimization strategies that combine full and stochastic gradients with additional techniques such as noise injection, dynamic batch sizing, and advanced sampling methods. These approaches have been shown to achieve favorable convergence rates and improved accuracy in practice~\citep{Zhang2012cimmuncation, shalev2013stochasticdual, Johnson2013accelerating, Defazio2014saga, Arjevani2015complexityl, Lin2015universal, Allen2017firstdirect}.

With the rapid advancement of optimization theory over the past several decades, significant research efforts have focused on nonsmooth nonconvex optimization problems. Notable approaches include gradient sampling methods~\citep{burke2002approximating, burke2005robust, kiwiel2007convergence, burke2020gradient}, bundle methods \citep{kiwiel1996restricted, atenas2023unified, gaudioso2023bundle}, and subgradient methods \citep{benaim2005stochastic, daniilidis2020pathological, bolte2021conservative}. In particular, Burke et al. developed several gradient-based optimizers for nonsmooth nonconvex objective functions by sampling gradients at a set of randomly generated nearby points~\citep{burke2002approximating, burke2005robust}. These sampled gradients are then used to construct a local search direction that approximates the $\epsilon$-steepest descent direction, where $\epsilon$ denotes the sampling radius~\citep{burke2020gradient}. Further contributions by Daniilidis and Drusvyatskiy examined the limitations of traditional subgradient methods in nonsmooth nonconvex settings. They demonstrated that the standard subgradient method may fail to converge to any Clarke stationary point, even when applied to Lipschitz-continuous functions~\citep{daniilidis2020pathological}. Moreover, recent work has introduced a randomized variant of Goldstein’s subgradient method, establishing a dimension-independent complexity bound for finding a $(\delta,\epsilon)$-Goldstein stationary point. This work builds on foundational concepts from Clarke’s generalized gradients~\citep{clarke1974necessary, clarke1975generalized, clarke1981generalized, clarke1990optimization, jordan2023deterministic}.

Despite significant progress with state-of-the-art optimizers, efficiently approximating the global optimum of nonsmooth nonconvex objective functions remains a major challenge~\citep{jordan2023deterministic}. Even advanced methods such as ADAM~\citep{Kingma2014adam} and stochastic variance reduced gradient (SVRG)~\citep{lian2017finite, nan2023extragradient, dubois2022svrg} often struggle with these complex landscapes, particularly in high-dimensional settings. A notable example arises in the context of deep neural networks (DNNs), whose objective functions are typically nonsmooth and nonconvex, and only Lipschitz continuous (see Definition~\ref{def1})~\citep{jordan2023deterministic}. This is largely due to the depth of DNNs and the use of non-differentiable activation functions, such as rectified linear units (ReLUs)~\citep{lecun2015deep, nair2010rectified, jordan2023deterministic}.

Importantly, although it has been demonstrated that deterministic, dimension-free, first-order optimizers cannot outperform randomized methods in smooth or convex regimes~\citep{clarke1990optimization, nesterov2005smooth, nesterov2018lectures, arjevani2023lower, jordan2023deterministic}, the complexity of nonsmooth nonconvex problems suggests that randomization could play a critical role in improving optimization outcomes. Therefore, developing a theoretical understanding of why and how randomization enhances the performance of gradient-based optimizers in these challenging settings is both timely and necessary~\citep{jordan2023deterministic}. In this work, we aim to contribute to this understanding by analyzing the efficacy of permutation randomization in nonsmooth nonconvex optimization and validating our theoretical findings through comprehensive numerical experiments.

\textit{Our Contributions}: This work presents comprehensive theoretical and experimental studies to uncover the advancements of permutation randomization, providing a partial answer to the fundamental question raised in~\citep{jordan2023deterministic}. Our key contributions are summarized as follows:

1. Limitation of Gradient-Based Optimizers without Randomization (Theorem~\ref{thm22}): Considering a gradient-based optimizer as an operator (see Definition \ref{def2})~\citep{Rudin1973functionalanalysis, yosida2012functional}, we show that such optimizers usually converge to a fixed point or a stationary point, e.g., a Clarke or $(\delta, \epsilon)$-Goldstein stationary point. Crucially, without randomization, the optimizer cannot guarantee to approximate a closed cube of the global optimum (Definitions~\ref{def5} and~\ref{def6}).

2. Advancement in Permutation Randomization (Theorem~\ref{thm31}): We theoretically establish that incorporating permutation randomization enables collectively covering the entire domain of the objective function over sufficiently large iterations. This property makes it feasible to approximate a closed cube (Definition~\ref{def6}) that contains the global optimum.

3. Permutation Randomization Preserves Convergence Rate (Theorem~\ref{thm32}): A natural concern is whether permutation randomization adversely affects the convergence rate of the underlying optimizer. We demonstrate that it does not: the norm of the permutation operator (Definition~\ref{def3}) is 1, implying that it preserves the convergence rate of the incorporated optimizer. Moreover, we provide a theoretical upper bound for this convergence rate in Theorem~\ref{thm32}.

4. Experimental Validation: Various numerical experiments demonstrate that integrating permutation randomization into ADAM~\citep{Kingma2014adam} consistently improves accuracy compared to three peer optimizers. Additionally, in noisy optimization tasks, ADAM with permutation randomization outperforms all baselines, further validating our theoretical insights.

Full proofs and extended analysis can be found in Appendices A and B.

\section{Preliminaries}
\label{gen_inst}

This section introduces the fundamental optimization problem investigated in this work, along with the key definitions and assumptions that underpin the subsequent analyses. We consider the following minimization problem:
\begin{equation} \label{eq1}
\begin{gathered}
\min_{x \subseteq \mathbb{R}^{D}}f(x)
\end{gathered}
\end{equation}

We assume that the real objective function $f:\mathbb{R}^{D} \rightarrow \mathbb{R}$ is Lipschitz continuous (please refer to Definition~\ref{def1}) over a closed set $I$ and $x=\lbrace x_1,x_2,\cdots,x_D \rbrace \subseteq I$. In addition, we define a closed subset $I^{\prime} \subseteq I$, within which $f$ is convex \slash smooth, while it is nonsmooth and nonconvex on the remaining $ I \backslash I^{\prime}$. To support the theoretical analyses that follow, we first introduce a set of mathematical symbols used throughout this work:

\begin{table}[htbp]
  \caption{The definitions of mathematical symbols}
  \centering
  \begin{tabular}{cc}
    \toprule               \\
  Symbols     &  Definitions  \\
    \midrule
      $\mathcal{G}$ & The gradient operator       \\
      $\mathcal{G}^t$ & The gradient operator in $t$ iterations \\
      $\mathcal{R}$ &   The randomization operator    \\
      $\mathcal{R}^t$ &   The randomization operator in $t$ iterations \\
      $I$ &  The domain of an objective function         \\
      $I_0$ & An initial set \\
      $T$ & The maximum iteration \\
      $B(x,\delta)$ & A closed cube with center $x$ and radius $\delta>0$ \\
      $D$ & Dimensionality of the objective function variable \\
    \bottomrule
  \end{tabular}
  \label{table:table1}
\end{table}

Furthermore, the following assumptions and definitions serve as essential prerequisites for the theoretical analyses of permutation randomization in nonsmooth nonconvex optimization problems.

\begin{assumption}
\label{assumption1}
For a set of points $\lbrace x_1, x_2, \cdots, x_D \rbrace \subseteq I$, representing all dimensions of the objective function variable, we assume the existence of a closed cube that covers each point. Specifically, for each coordinate $x_i$, there exists a closed cube $B(x_i, \delta_i), 1 \leq i \leq D, \delta_i>0$, such that $x_1 \in B(x_1,\delta_1), x_2 \in B(x_2,\delta_2), \cdots, x_D \in B(x_D, \delta_D)$. Notably, $x_i$ is the center of closed cube $B(x_i,\delta_i)$ and  $\delta_i$ denotes the radius.
\end{assumption}

\begin{definition}
\label{def1}
(\textit{Lipschitz Continuous Function}) For any $x, y \subseteq \mathbb{R}^{D}$, $\left \| f(x) - f(y) \right \| \leq L \left \| x - y \right \|, L \in \mathbb{R}$ and $L < \infty$, we call $f$ Lipschitz continuity and can rewrite as $f \in Lip(I)$.
\end{definition}

\begin{definition}
\label{def2}
(\textit{Gradient-based Optimization Operator}) To conveniently employ Real Analysis and Functional Analysis for theoretical analyses, we denote a gradient-based optimizer without randomization as an operator:  $\mathcal{G}: \mathbb{R} \rightarrow \mathbb{R}^D$. $\mathcal{G}$ can represent any gradient-based optimizer without randomization, such as gradient descent (GD)~\citep{von2023transformers}, Nesterov accelerated gradient descent~\citep{lin2019nesterov}, and ADAM~\citep{Kingma2014adam}. For instance, to optimize objective function $f$ using GD, if $x \in \mathbb{R}$, we have $\mathcal{G} \cdot f(x) = x - \alpha \cdot \nabla f(x)$, and $\alpha$ denotes the step size of GD \citep{von2023transformers}.
\end{definition}

\begin{definition}
\label{def3}
(\textit{Permutation Randomization Operator}) We define permutation randomization as an operator: $\mathcal{R}:\mathbb{R}^D \rightarrow \mathbb{R}^D, D \in \mathbb{N}, D>1$. Specifically, when applied to a vector in three dimensions, permutation randomization reorders all components. For example: $\mathcal{R} \cdot \lbrace x_1, x_2, x_3 \rbrace \rightarrow \lbrace x_3, x_1, x_2 \rbrace$.
\end{definition}

\begin{definition}
\label{def4}
(\textit{An Initial Set}) We initialize a closed interval $I_0 = \lbrace x_{1,0}, x_{2,0}, \cdots, x_{D,0} \rbrace \subseteq I$ for the gradient-based optimization operator $\mathcal{G}$, and begin optimizing the objective function $f$. For example, let $D=3$ and $I_0 = \lbrace 10, 20, 60 \rbrace$ be the initial set for the gradient-based optimizer.
\end{definition}

\begin{definition}
\label{def5}
(\textit{The Iterative Format of Gradient-based Optimization and Permutation Randomization Operator}) Suppose $t$ is the current iteration, we denote the iterative format of the gradient-based optimization and permutation randomization operator as $\mathcal{G}^t$ and $\mathcal{R}^t$, respectively. Naturally, we can represent the output of a gradient-based optimizer in iteration $t$ as a closed interval $\mathcal{G} ^t\cdot f(I_0) \rightarrow \lbrace x_{1,t}, x_{2,t}, \cdots, x_{D,t} \rbrace$. Similarly, the output of randomization applied to a set of points represents $\mathcal{R}^t \cdot  I_0  \rightarrow  \lbrace \hat{x}_{1,t},\hat{x}_{2,t},\cdots, \hat{x}_{D,t} \rbrace$. For example, $\mathcal{G}^2 \cdot f(I_0) = \mathcal{G} \cdot \mathcal{G} \cdot f(I_0) = \lbrace x_{1,2},x_{2,2}, \cdots, x_{D,2} \rbrace$ and $\mathcal{R}^2 \cdot \mathcal{G}^2 \cdot f(I_0) = \mathcal{R} \cdot \mathcal{R} \cdot \mathcal{G}^2 \cdot f(I_0) \cdot \lbrace x_{1,2},x_{2,2}, \cdots, x_{D,2} \rbrace = \lbrace \hat{x}_{1,2},\hat{x}_{2,2},\cdots, \hat{x}_{D,2} \rbrace$.
\end{definition}

\begin{definition}
\label{def6}
(\textit{The Closed Cubes of Global Optimum}) The closed cubes of the global optimum are represented as a series of closed cubes as $B(x_{1,gbest},\delta_1), B(x_{2,gbest},\delta_2), \cdots, B(x_{D,gbest},\delta_D) \subseteq I$.
\end{definition}

\begin{definition}
\label{def8}
(\textit{A Class of Closed Cubes Cover The Gradient}) According to Definition~\ref{def5} and Assumption~\ref{assumption1}, any gradient-based optimizer incorporating permutation randomization $\mathcal{R}$ applied on $ f(I_0)$ in iterations $t$ is denoted as $\mathcal{R}^t \cdot \mathcal{G}^t \cdot f(I_0) = \lbrace \hat{x}_{1,t}, \hat{x}_{2,t}, \cdots, \hat{x}_{D,t} \rbrace$. Naturally, a set of closed cubes that cover these points can be denoted as $B(\hat{x}_{1,t},\delta_{1,t}), B(\hat{x}_{2,t},\delta_{2,t}), \cdots, B(\hat{x}_{D,t},\delta_{D,t})$.
\end{definition}

\section{The Limitation of Conventional Gradient-based Optimizers}

\begin{lemma}
\label{lem21}
(Contraction Property of Gradient-based Optimizer without Randomization) According to Definitions~\ref{def2}, ~\ref{def3}, ~\ref{def4}, and~\ref{def5}, a gradient-based optimizer without randomization $\mathcal{G}:\mathbb{R} \rightarrow \mathbb{R}^D$ within iterations $t$ denoted as $\mathcal{G} \cdot f(I_0) = \lbrace x_{1,t}, x_{2,t}, \cdots, x_{D,t} \rbrace$, $\left \| \mathcal{G}^{t+1} \right \| \leq \left \| \mathcal{G}^{t} \right \|,\forall t \in \lbrack 1,T \rbrack$ holds.
\end{lemma}

\begin{theorem}
\label{thm22}
(Limitation of Gradient-based Optimizer without Randomization) According to Definitions~\ref{def2}, ~\ref{def3}, ~\ref{def4}, ~\ref{def5}, and~\ref{def6}, suppose current iteration as $t$, maximum iteration as $T$, and initialized interval as $I_0$, if a gradient-based optimizer without randomization $\mathcal{G}$ can approximate the closed cubes of global optimum $B(x_{i,gbest},\delta_i), 1 \leq i \leq D, \delta_i > 0$,  for all closed cubes of global optimum $B(x_{1,gbest},\delta_1),B(x_{2,gbest},\delta_2), \cdots, B(x_{D,gbest},\delta_D)  \subseteq \mathcal{G}^t \cdot f(I_0)$ must hold.
\end{theorem}

Theorem \ref{thm22} showcases a rigorous prerequisite for any gradient-based optimizer without randomization to approximate a closed cube of global optimum within maximum iteration $T$. It demonstrates that the gradient-based optimizer on nonsmooth nonconvex problems without randomization gradually narrows its sight and gets stuck in the nearest stationary point, such as $(\delta,\epsilon)$-Goldstein stationary point~\citep{clarke1974necessary, clarke1975generalized, clarke1981generalized, clarke1990optimization}, after iterations $t$. The proof of Lemma \ref{lem21} and Theorem \ref{thm22} can be viewed in Appendix A, Supplementary Material.

\section{Theoretical Analyses on Permutation Randomization}
In this section, we theoretically analyze the advantages of permutation randomization. Unlike the gradient-based optimizer without randomization, the gradient-based optimizer incorporating permutation randomization guarantees to approximate the closed cube of global optimum, only considering that the maximum iteration $T$ is sufficiently large. 

\begin{figure}[H]
  \centering
  \includegraphics[width=0.76\textwidth]{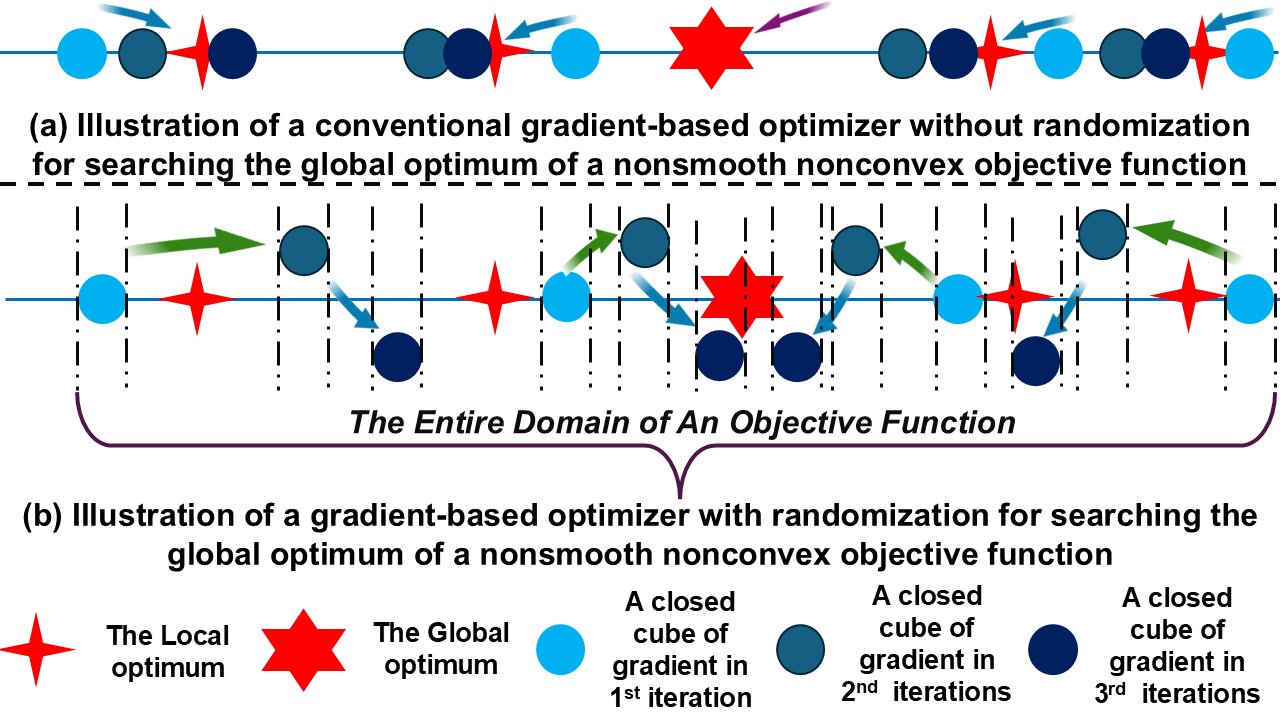}
  \caption{An illustration of a gradient-based optimizer incorporating permutation randomization can cover a closed cube of global optimum when the maximum iteration $T$ is sufficiently large.}
  \label{fig:fig1}
\end{figure}

Figure~\ref{fig:fig1} illustrates the contrasting outcomes of applying a gradient-based optimizer with and without coordinate randomization. Subfigure (a) demonstrates that a conventional gradient-based optimizer often becomes trapped in multiple local optima, primarily due to the vanishing gradient phenomenon near these points. In contrast, subfigure (b) depicts the behavior of a gradient-based optimizer enhanced with coordinate randomization. One of the key advantages of permutation randomization is its capability to disrupt the contraction or shrinkage property of conventional gradient-based optimizers by reordering all coordinates. This reordering provides an opportunity to escape local optima and potentially cover the entire domain of the objective function to approach the global optimum. In addition, the following Lemmas~\ref{lem31} and~\ref{lem32} are fundamental to our theoretical analyses employing real analysis~\citep{Royden1968realanalysis} and functional analysis~\citep{Rudin1973functionalanalysis}. These Lemmas are representative theorems in set theory~\citep{Royden1968realanalysis}.

\begin{lemma}
\label{lem31}
\textbf{(Heine-Borel Theorem)}. ~\citep{Royden1968realanalysis} Assume $\Gamma$ is a close and bounded set and $\lbrace g_i \rbrace_{i=1}^K=G$ is a closed set. Then $ \bigcup_{i=1}^K g_i \supseteq \Gamma,$ and $\overline{\overline{G}} =\aleph_0$.
\end{lemma}

\begin{lemma}
\label{lem32}
\textbf{(Vitali Covering Theorem)}. ~\citep{Royden1968realanalysis} Assume $\lbrace B_i \rbrace_{i=1}^n$ are closed sets and $\forall B_i \cap B_j = \emptyset$, $i \neq j$, $\Gamma \subseteq \mathbb{R}$, and $m^\star (\Gamma) < \infty$, if $m^\star(\Gamma \backslash \bigcup_{i=1}^n B_i) < \epsilon$, $\forall \epsilon >0$, holds, $\lbrace B_i \rbrace_{i=1}^n$ defines a Vitali Covering of $\Gamma$. And $m^\star( \cdot )$ denotes a Lebesgue outer measure.
\end{lemma}

Importantly, we present Theorem \ref{thm31} as follows to demonstrate that an optimizer incorporating permutation randomization guarantees that all output intervals such as $\mathcal{R}^t \cdot \mathcal{G}^t \cdot f({I_0}) \rightarrow \lbrack \hat{x}_t,\hat{y}_t \rbrack$ (please refer to Definition \ref{def8}) can cover a closed cube of the global optimum.

\begin{theorem}
\label{thm31}
(The Advantage of Gradient-based Optimizer Incorporating Permutation Randomization) According to Definitions~\ref{def1}, ~\ref{def2}, ~\ref{def3}, ~\ref{def4}, and~\ref{def6}, given a Lipschitz continuous but nonsmooth nonconvex objective function $f$ (please refer to Definition~\ref{def1}), an optimizer without randomization $\mathcal{G}$ (please refer to Definition~\ref{def2}), a randomization operator $\mathcal{R}$ (please refer to Definition \ref{def3}), an initialized set $I_0=\lbrace x_{1,0}, x_{2,0}, \cdots, x_{D,0} \rbrace$ (please refer to Definition~\ref{def4}), the close cubes of the global optimum are denoted as $B(x_{1,gbest},\delta_1), B(x_{2,gbest},\delta_2), \cdots, B(x_{D,gbest},\delta_D)$ (please refer to Definition~\ref{def6}, suppose the maximum iteration is $T$, the following 1), 2), and 3) hold:

1) According to Definition \ref{def8}, $\mathcal{R}^t \cdot \mathcal{G}^t \cdot f(I_0) = \lbrace \hat{x}_{1,t},\hat{x}_{2,t}, \cdots, \hat{x}_{D,t} \rbrace \subseteq I$ holds. According to Assumption \ref{assumption1} and $D$, $T$ are sufficiently large, there are $B(\hat{x}_{k,i},\delta_{k,i}) \cap B(\hat{x}_{k,j},\delta_{k,j}) \neq \emptyset, 1<i,j<T$, inferring from Lemma \ref{lem31}:

\begin{equation} \label{eq2}
\begin{gathered}
 \forall B(x_{i,gbest},\delta_i) \subseteq I \subseteq \bigcup_{t=1}^T \bigcup_{i=1}^D  B(\hat{x}_{i,t},\delta_{i,t}), 1 \leq i \leq D, 1 \leq t \leq T \\
\end{gathered}
\end{equation}

2) According to Definition~\ref{def8}, $\mathcal{R}^t \cdot \mathcal{G}^t \cdot f(I_0) = \lbrace \hat{x}_{1,t},\hat{x}_{2,t}, \cdots, \hat{x}_{D,t} \rbrace \subseteq I$ holds. According to Assumption \ref{assumption1}, and $D$, $T$ are sufficiently large, there are $B(\hat{x}_{k,i},\delta_{k,i}) \cap B(\hat{x}_{k,j},\delta_{k,j}) = \emptyset, 1<i,j<T$, inferring from Lemma \ref{lem32}:

\begin{equation} \label{eq3}
\begin{gathered}
\forall B(x_{i,gbest},\delta_i) \subseteq I \subseteq \bigcup_{t=1}^T \bigcup_{i=1}^D  B(\hat{x}_{i,t},\delta_{i,t}) \cup E, 1 \leq i \leq D, 1 \leq t \leq T\\
\overline{\overline{\textit{E}}}= \aleph_0
\end{gathered}
\end{equation}

3) Otherwise, if some closed cubes are overlapped, and others are not, it needs to consider 1) and 2) comprehensively. In brief, the randomized intervals generated by any optimizer can cover the closed cubes of the global optimum.
\end{theorem}

In Theorem~\ref{thm31}, it concludes that an set of closed cubes of the global optimum $B(x_{i,gbest},\delta_i), 1 \leq i \leq D, \delta_i > 0$ (please refer to Definition~\ref{def6}) can be covered in the union of a series closed cubes, such as $B(\hat{x}_{i,t},\delta_{i,t}), 1 \leq i \leq D, 1 \leq t \leq T$, generated via the gradient-based optimizer incorporating randomization, given sufficiently large iterations. For instance, optimize an objective function $f$ with an initialized interval $I_0=\lbrace 10, 11, 20 \rbrace$ and the global best optimum $I_{gbest} = \lbrace 19, 16, 15 \rbrace$. Suppose we employ a gradient-based optimizer $\mathcal{G}$, due to contraction property of gradient-based optimizer, we have $\mathcal{G} \cdot f(I_0) = \lbrace 12, 13, 14 \rbrace$, $\mathcal{G}^2 \cdot f(I_0) = \lbrace 14, 15, 16 \rbrace$, and $\mathcal{G}^3 \cdot f(I_0) = \lbrace 15, 16, 19 \rbrace$. If we assume the global optimum is $\lbrace 19, 16, 15 \rbrace$, the last randomization can facilitate an accurate approximation to the global optimum, such as $\mathcal{R} \cdot \lbrace 15, 16, 19 \rbrace \rightarrow \lbrace 19, 16, 15 \rbrace$. Meanwhile, we can derive the global optimum $\lbrace 19, 16, 15 \rbrace \subseteq \lbrace 12, 13, 14\rbrace \cup \lbrace 14, 15, 16 \rbrace \cup \lbrace 15, 16, 19\rbrace$.

Furthermore, as $D$ becomes sufficiently large (e.g., $D \rightarrow \infty$), Eqs.~\eqref{eq2} and~\eqref{eq3} incorporate an increasing number of gradient cubes to effectively cover the entire domain of the objective function.

Moreover, we provide Lemma~\ref{lem33} and Theorem~\ref{thm32} to analyze the potential influence of permutation randomization on the convergence rate. In detail, supposing the permutation randomization as an operator (please refer to Definition~\ref{def3}, we can prove that the norm of $\mathcal{R}$ is equal to 1 and discuss the upper bound of the gradient-based optimizer incorporating randomization.

\begin{lemma}
\label{lem33}
(Norm of Permutation Randomization Operator is Equal to 1) Suppose the permutation randomization as an operator $\mathcal{R}:\mathbb{R}^D \rightarrow \mathbb{R}^D$, $\left \| \mathcal{R} \right \| =1$ holds, if $D < \infty$.
\end{lemma}

It is not difficult to prove Lemma \ref{lem33}. According to Definition \ref{def3}, we can infer:
\begin{equation} \label{eq4}
\begin{gathered}
\left \| \mathcal{R} \cdot \lbrace x_1, x_2, \cdots, x_D \rbrace 
 \right \|
 =  \left \| \lbrace \hat{x}_1, \hat{x}_2, \cdots, \hat{x}_D \rbrace \right \|
\end{gathered}
\end{equation}

According to  Definition \ref{def5}, Lemma \ref{lem33}, and finite-dimensionality space, for any $X,Y \in I$, we have:
\begin{equation} \label{eq5}
\begin{gathered}
\left \| \mathcal{R}^t \cdot \mathcal{G}^t \cdot (f(X) -f(Y)) \right \| \leq \left \| \mathcal{R}^t \right \| \cdot \left \| \mathcal{G}^t \cdot (f(X) -f(Y)) \right \| =  \left \| \mathcal{G}^t  \cdot (f(X) -f(Y)) \right \|
\end{gathered}
\end{equation}

Let $X$ be $I_1=\mathcal{G} \cdot f(I_0)$ and $Y$ be $I_0$, inferring from Eq. \eqref{eq5}, we have:
\begin{equation} \label{eq6}
\begin{gathered}
\left \| \mathcal{R}^t \cdot \mathcal{G}^t \cdot (f(I_1) -f(I_0)) \right \| \leq \  \left \| \mathcal{G}^{t+1}  \cdot f(I_0) - \mathcal{G}^{t}  \cdot f(I_0)) \right \|
\end{gathered}
\end{equation}

The upper bound on the convergence rate of gradient-based optimizer~\citep{wang2023parallelization} incorporating randomization can reach $\left \| \mathcal{G}^{t+1}  \cdot f(I_0) - \mathcal{G}^{t}  \cdot f(I_0)) \right \|$; thus, we revealed the upper bound of the convergence rate of a gradient-based optimizer incorporating permutation randomization should be equivalent to the convergence rate of original gradient-based optimizer and proved that permutation randomization could maintain the convergence rate of incorporated optimizer and shown in Theorem~\ref{thm32}.

\begin{theorem}
\label{thm32}
(Permutation Randomization Preserves the Convergence Rate of the Incorporated Gradient-Based Optimizer) According to Definition \ref{def8}, the convergence rate of a gradient-based optimizer incorporating randomization should be smaller or equal to the convergence rate of the incorporated gradient-based optimizer $\mathcal{G}$ without randomization.
\end{theorem}

Suppose incorporating randomization with a smaller norm (e.g., $\left \| \mathcal{R} \right \| <1$), according to Eq.~\eqref{eq6}, it might decrease the convergence rate of the original optimizer. For instance, Burke's work \citep{burke2020gradient} demonstrates that randomization with a continuously reduced norm should benefit local searches but influence convergence. Furthermore, the following Theorem discusses the upper bound on the convergence rate of a gradient-based optimizer incorporating permutation randomization to approximate a closed cube of the global optimum.

\begin{theorem}
\label{thm33}
(Upper Bound on Convergence Rate of Randomized Gradient-based Optimizer Approximates a Closed Cube of Global Optimum) According to Definitions~\ref{def5}, ~\ref{def6}, and Theorem~\ref{thm32}, considering the objective function $\psi:\mathbb{R}^1 \rightarrow \mathbb{R}^1$, if maximum convergence rate is less than $ 2 \cdot \left| \psi(B(x_{gbest},\delta)) \right|$, $\lbrace x, x_{gbest} \in I \subseteq \mathbb{R}^D, \psi(x) \leq \psi(B(x_{gbest},\delta)) \rbrace  = \bigcup_{k=1}^T \bigcup_{m=1}^T \bigcap_{t=m}^T \lbrace x \in I \subseteq \mathbb{R}^D: max \lbrack \mathcal{R}^t \cdot \mathcal{G}^t \cdot \psi(x) \rbrack + \frac{1}{k}\rbrace$ holds.
\end{theorem}

Our theoretical studies indicate the efficacy of permutation randomization on nonsmooth nonconvex optimization. The following experimental studies can validate theoretical analyses and further demonstrate the advantages of permutation randomization.

\section{Experimental Results}
In this section, we empirically investigate the performance of permutation randomization with a gradient-based optimizer. Specifically, we incorporate the permutation randomization and a representative gradient-based optimizer ADAM~\citep{Kingma2014adam} as randomized ADAM (see pseudo-code in Table~\ref{table:table3}, Appendix C, Supplementary Material) to validate reconstruction loss, such as $\left \| InputMatrix - WeightMatrix \times FeatureMatrix \right \|_F^2 \slash \left \| InputMatrix \right \|_F^2$, via multiple numerical experiments on unsupervised deep nonlinear learning methods such as Deep Nonlinear Matrix Factorization (DNMF)~\citep{Wen2012lowrank, Shen2012lowrank, trigeorgis2014deep, trigeorgis2016deep} and Deep Belief Network (DBN)~\citep{hinton2009deep}, and noisy problems. Meanwhile, ADMM~\citep{Nishihara2015admm}, ADAM \citep{Kingma2014adam}, and SVRG~\citep{lian2017finite,dubois2022svrg, nan2023extragradient} are included as peer optimizers to validate permutation randomization.

Importantly, permutation randomization is only applied when the difference between the current and previous gradient reaches the given threshold. The threshold measuring the difference between the lastious and current gradient is $1.0 \times 10^{-2}$. We also expect to investigate that permutation randomization enables continuous update of the gradient during the late stage~\citep{clarke1974necessary, clarke1981generalized, goldstein1977optimization}.

Furthermore, we introduce the objective functions of DNMF, presented as follows:
\begin{subequations} \label{eq9}
\begin{align}
\min_{Z_i \in \mathbb{R}^{m \times n}} \bigcup_{i=1}^k \left \| Z_i\right \|_1 \label{eq9a}\\
\textit{s.t.} \quad (\prod_{i=1}^k X_i) \cdot \mathcal{N}_k (Y_k) + Z_k=S \label{eq9b}
\end{align} 
\end{subequations}
Optimizing the objective function of DNMF is more challenging than most regression problems \citep{trigeorgis2014deep, trigeorgis2016deep} since more variables (e.g., three variables) are involved in the optimization. In detail, $X_i$ represents the weight/mixing matrix at $i^{th}$ layer, $Y_k$ denotes the feature matrix at $k^{th}$ layer, $Z_k$ represents the noise/background components at $k^{th}$ layer and $S$ defines the input data matrix. In addition, $\mathcal{N}_k$ represents the activation function in $k^{th}$ layer. Notably, ReLU is determined as an activation function \citep{nair2010rectified}. The reconstruction loss is defined as $\frac{\left \| (\prod_{i=1}^k X_i) \cdot  Y_k  + Z_k-S \right \|_F^2}{\left \| S \right \|_F^2}$~\citep{trigeorgis2014deep, trigeorgis2016deep}. 

In addition, randomized ADAM and three other peer optimizers are validated on public biomedical data on \textit{OpenfMRI}~\citep{MBMEfMRI}. We employed a data augmentation technique~\citep{pei2022data} to expand the original dataset from 29 subjects to a total of 100 subjects. In this experimental study, only the maximum iterations are manually determined as two hundred iterations, and other parameters are tuned as default settings in~\citep{Lei2017SCSG, Cutkosky2019Momentum, Nishihara2015admm}. Meanwhile, the experimental studies are validated on the CPU cluster( Intel Xeon Gold 6246R) and NVIDIA CUDA cores (NVIDIA DGX A100 P3687).

Moreover, the following figures (from Figure~\ref{fig:fig2} to~\ref{fig:fig4} present all experimental results within 2000 iterations. All figures provide the reconstruction loss of the subjects' signal matrices. Overall, randomized ADAM and three other peer optimizers (ADMM~\citep{Nishihara2015admm}, ADAM~\citep{Kingma2014adam}, and SVRG~\citep{lian2017finite}) are applied to optimize the objective function of DNMF (please refer to Eqs.~\eqref{eq9a} and~\eqref{eq9b}. In addition, with an automatic estimation technique provided in~\citep{Shen2012lowrank, Wen2012lowrank}, all subjects' signal matrices can be consistently decomposed into two layers. Notably, in Figures ~\ref{fig:fig2} and ~\ref{fig:fig3}, we provide randomly selected four subjects' reconstruction accuracy curves of the first and second layers. 
\begin{figure}[htbp]
  \centering
  \includegraphics[width=0.70\textwidth]{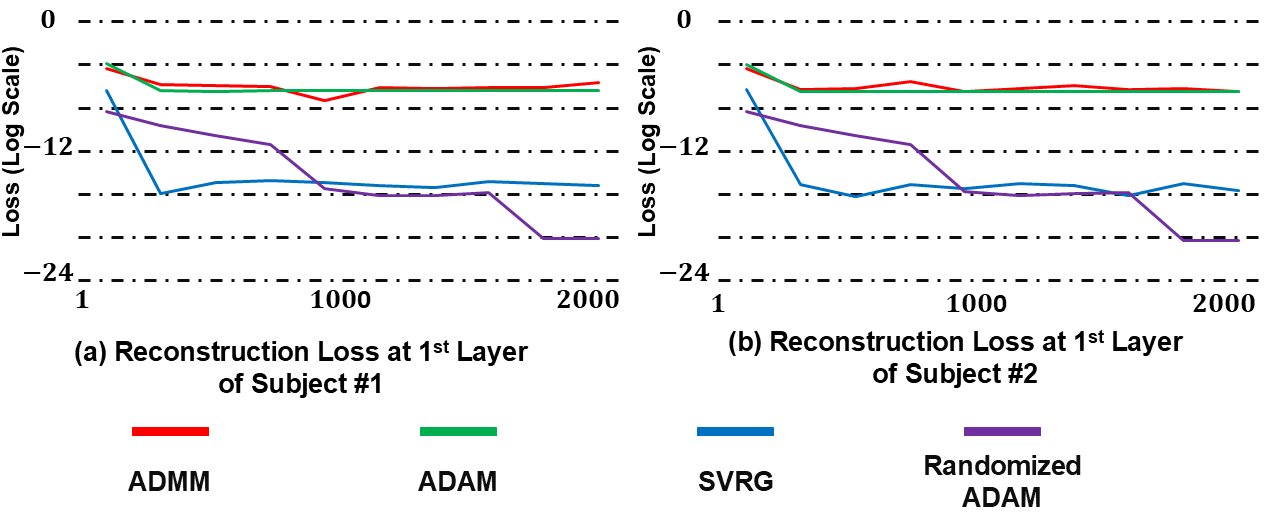}
  \caption{The reconstruction loss comparison of randomized ADAM and the other three peer optimizers within two thousand iterations of randomly selected two subjects at the first layer.}
  \label{fig:fig2}
\end{figure}

\begin{figure}[htbp]
  \centering
  \includegraphics[width=0.70\textwidth]{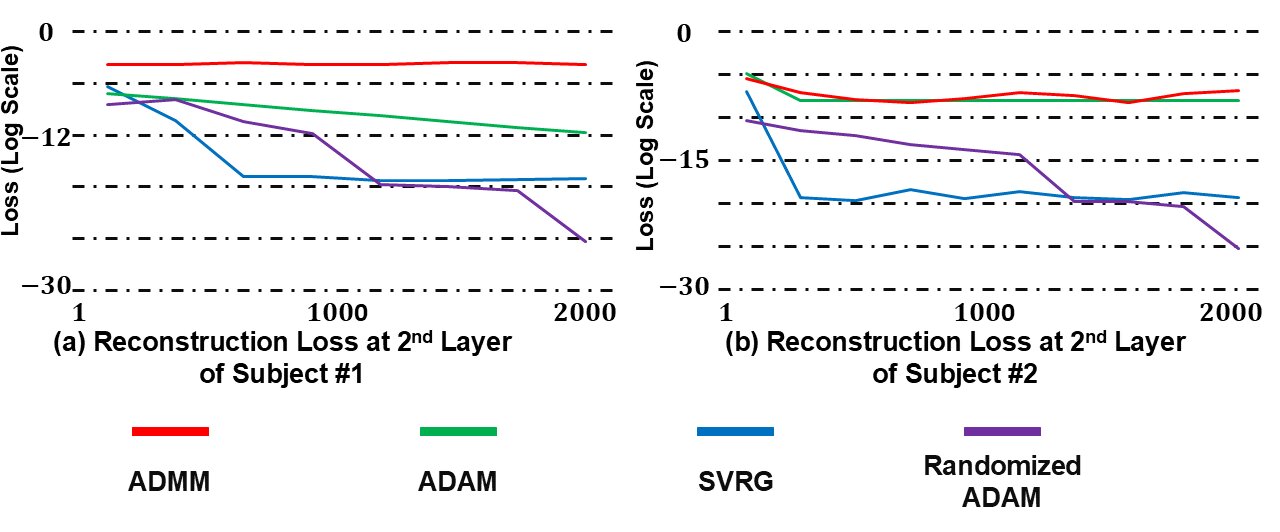}
  \caption{The reconstruction loss comparison of randomized ADAM and the other three peer optimizers within two thousand iterations of randomly selected two subjects at the second layer.}
  \label{fig:fig3}
\end{figure}
According to Figures~\ref{fig:fig2},~\ref{fig:fig3}, and~\ref{fig:fig4}, ADAM incorporating permutation randomization can achieve the highest reconstruction accuracy. Overall, SVRG and randomized ADAM perform better than ADMM and ADAM. Notably, the gradient of randomized ADAM can continuously update even in the late stage, after 1600 iterations in Figure~\ref{fig:fig2} (a) and (b). Specifically, when the number of iterations is smaller than 1000, SVRG can achieve a higher accuracy than randomized ADAM. In contrast, if the number of iterations is larger than 100, randomized ADAM achieves the highest reconstruction accuracy. In addition, to evaluate the performance of ADAM incorporating permutation randomization on all subjects, we also provide the averaged reconstruction loss in Figure~\ref{fig:fig3} that can further compare the performance of randomized ADAM with other peer algorithms.

Meanwhile, we aggregate all subjects as group-wise data~\citep{MBMEfMRI} and employ a 3-layer DBN to further validate the performance of randomized ADAM and the other two peer optimizers. We also introduce a noisy optimization problem to investigate the performance of permutation randomization on nonsmooth nonconvex problems~\citep{arjevani2023lower, forneron2023noisy, lei2023stability}.
\begin{figure}[htbp]
  \centering
  \includegraphics[width=0.70\textwidth]{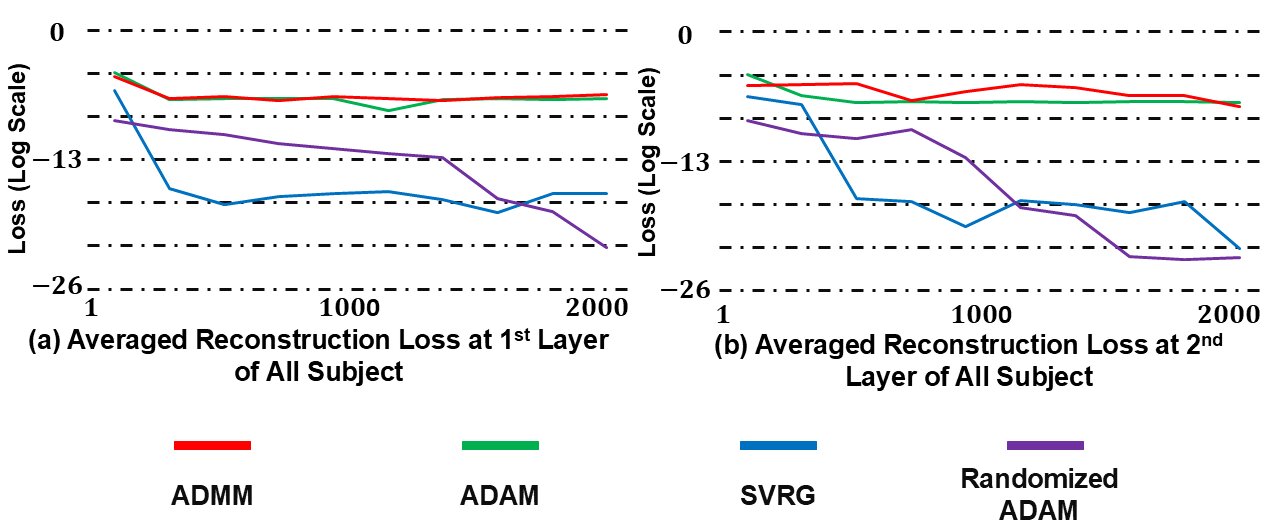}
  \caption{The averaged reconstruction loss comparison of randomized ADAM and the other three peer optimizers within two thousand iterations across all subjects at first and second layers, respectively.}
  \label{fig:fig4}
\end{figure}
Consistent with earlier numerical experiments, the randomized gradient-based optimizer is less prone to becoming trapped in stationary points in both DNNs and noisy DNMF (introducing random noise into the gradient at every 10 iterations), as shown in Figures~\ref{fig:fig6} and~\ref{fig:fig7} in Appendix C of the Supplementary Materials. For additional evaluation, we applied each method using default parameter settings~\citep{Kingma2014adam, Allen2012featuredetectors, Lei2017SCSG, lei2023stability} to a logistic regression task~\citep{schober2021logistic} using the publicly available breast cancer dataset~\citep{BreastCancer}, visualized in Figure~\ref{fig:fig8} in Appendix C of the Supplementary Material.

\section{Statistical Analyses}
\begin{figure}[htbp]
  \centering
  \includegraphics[width=0.76\textwidth]{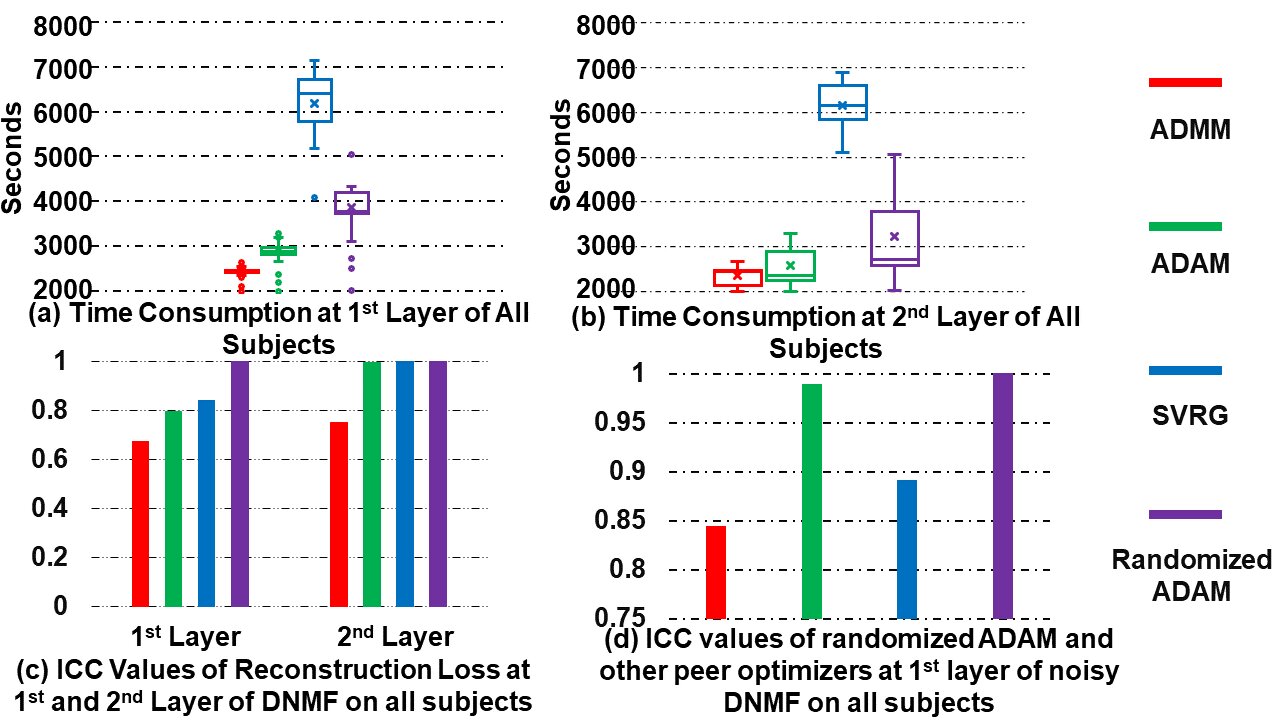}
  \caption{An illustration of time-consumption and consistency comparisons of randomized ADAM and other peer optimizers. The box plots in (a) and (b) represent the time consumption of four algorithms using all subjects; in Figure~\ref{fig:fig5}(c) provides the ICC values to demonstrate an overall consistency.}
  \label{fig:fig5}
\end{figure}

In this section, we present a quantitative analysis of the previously reported experimental results. Since all gradient-based optimizers examined in our empirical study are iterative algorithms, the reconstruction loss between adjacent iterations is not statistically independent (see Figures~\ref{fig:fig2}, ~\ref{fig:fig3}, and ~\ref{fig:fig4}). This lack of independence limits the applicability of traditional statistical tests such as t-tests and confidence intervals for comparing iterative reconstruction performance~\citep{field2013discovering}. To address this limitation, we adopt intra-class correlation coefficients (ICCs)—a widely used descriptive statistical technique for assessing consistency in grouped quantitative measurements~\citep{bartko1976various, bujang2017simplified}. Figure~\ref{fig:fig5}(c) and (d) report the ICCs for randomized ADAM and three peer optimizers. Notably, randomized ADAM and SVRG exhibit the highest levels of consistency. For example, at the second layer of DNMF, ICCs for randomized ADAM exceed 0.80 at the first layer and surpass 0.90 at the second layer, indicating strong internal reliability. These results highlight the performance of permutation randomization as comparable to SVRG, which uses averaged gradients for updates. In addition to consistency, we evaluated time consumption across all methods. As shown in Figures~\ref{fig:fig5}(a) and (b), the runtime of permutation randomization is higher than ADAM and ADMM, but substantially lower than SVRG, striking a promising balance between performance and efficiency. The mean and standard deviation of these results are summarized in Table~\ref{table:table2}.

\begin{table}[htbp]
  \caption{Time Consumption Comparison of All Optimizers in Seconds}
  \centering
  \begin{tabular}{cc}
    \toprule               \\
   Time Consumption at 1st Layer of DNMF     &  Time Consumption at 2nd Layer of DNMF  \\
    \midrule
    ADMM $451.58 \pm 93.86$     & ADMM $287.42 \pm 78.15$       \\
    ADAM  $901.65 \pm 196.96$   & ADAM  $579.98 \pm 53.41$      \\
    SVRG  $4538.50 \pm 305.49$  &  SVRG  $4441.60 \pm 547.12$         \\
    Randomized ADAM $1562.50 \pm 185.31$  & Randomized ADAM $1228.70 \pm 195.09$ \\
    \bottomrule
  \end{tabular}
  \label{table:table2}
\end{table}

\section{Conclusion}
In this work, we delivered theoretical and experimental analyses of permutation randomization in nonsmooth nonconvex optimization. Specifically, we showcased that incorporating permutation randomization into a gradient-based optimizer leads to covering various closed cubes that encompass the global optimum. Additionally, we indicated that this technique does not degrade the original optimizer’s convergence rate. Lastly, this work will inspire a wide array of theoretical and experimental investigations to explore the efficacy of other randomization techniques.

\bibliography{sadamnewbib}

\newpage

\appendix

\section{Appendix A}

\textbf{Lemma} 3.1 (Contraction Property of Gradient-based Optimizer without Randomization) According to Definitions~\ref{def2},~\ref{def3},~\ref{def4},~\ref{def5}, and Assumption~\ref{assumption1}, a gradient-based optimizer without randomization $\mathcal{G}:\mathbb{R} \rightarrow \mathbb{R}^D$ within $t$ iterations denoted as $\mathcal{G}^t \cdot f(I_0)$, $\forall t \in \lbrack 1,T \rbrack$, $\mathcal{G} \cdot f(I_t)= I_{t+1}$, $I_{t+1} \subseteq I_t$.\\
\textbf{\textit{Proof}}:\\
\textit{\textbf{Proof by contradiction}}.
Assume $\mathcal{G} \cdot f(I_t) = I_{t+1} \supseteq I_t$ and initialized closed interval as $I_0$, we can infer:
\begin{equation} \tag{A1} \label{eqA1}
\begin{gathered}
  I_T \supseteq \cdots \supseteq I_t  \supseteq \cdots \supseteq I_0
\end{gathered}
\end{equation}

Let $T \rightarrow \infty$, we have $ I_T \rightarrow (-\infty,+\infty)$ which implies:
\begin{equation} \tag{A2} \label{eqA2}
\begin{gathered}
  \vert \vert \mathcal{G}^t \cdot f(I_0) \vert \vert  \rightarrow \infty
\end{gathered}
\end{equation}

Based on the concept of a contraction operator, we infer that $\mathcal{G}$ is not a contraction and therefore cannot guarantee convergence. This supports our theoretical result that a gradient-based optimizer must exhibit contraction behavior, by inducing the shrinkage of input closed sets or intervals, to ensure convergence properties.

\textbf{Theorem} 3.2 (Limitation of Gradient-based Optimizer without Randomization) According to \textbf{Definitions}~\ref{def2},~\ref{def3},~\ref{def4},~\ref{def5}, and~\ref{def6}, suppose current iteration as $t$, maximum iteration as $T$, and initialized interval as $I_0$ if a gradient-based optimizer without randomization $\mathcal{G}$ can approximate the closed cubes of global optimum $B(x_{i,gbest}, \delta_i), \delta_i>0, 1 \leq i \leq D,$, for any $t \in \lbrack 1, T \rbrack$, $B(x_{i,gbest}, \delta_i) \subseteq I_t = \mathcal{G}^t \cdot f(I_0)$ must hold.\\

\textbf{\textit{Proof}}:\\
Inferring from Lemma \ref{lem21}, we have:
\begin{equation} \tag{A3} \label{eqA3}
\begin{gathered}
 \mathcal{G}^T \cdot f(I_0) \subseteq \cdots \subseteq  \mathcal{G}^t \cdot f(I_0) \subseteq \cdots \subseteq \mathcal{G}^1 \cdot f(I_0) 
\end{gathered}
\end{equation}

Let $t \rightarrow \infty$, consider $\mathcal{G}$ can approximate a closed cube of global optimum $B(x_{gbest},\delta), \delta >0$. And inferring from \textit{Cantor Intersection Theorem}, we have:
\begin{equation} \tag{A4} \label{eqA4}
\begin{gathered}
 \forall i \in \lbrack 1,D \rbrack, B(x_{i,gbest},\delta_i) \subseteq \bigcap_{t=1}^T \mathcal{G}^t \cdot f(I_0)
\end{gathered}
\end{equation}

Therefore, we conclude that for a conventional gradient-based optimizer to successfully approach the global optimum, the intervals generated in each iteration, denoted as $I_t, 1 \leq t \leq T$, must include the global optimum.

\section{Appendix B}

\textbf{Theorem} 4.3 (The Advantage of Gradient-based Optimizer Incorporating Permutation Randomization) According to Definitions \ref{def1}, \ref{def2}, \ref{def3}, \ref{def4}, and \ref{def6}, given a Lipschitz continuous but nonsmooth nonconvex objective function $f$ (please refer to Definition \ref{def1}), an optimizer without randomization $\mathcal{G}$ (please refer to Definition \ref{def2}), a randomization operator $\mathcal{R}$ (please refer to Definition \ref{def3}), an initialized internal $I_0=\lbrace x_{1,0}, x_{2,0}, \cdots, x_{D,0} \rbrace$ (please refer to Definition \ref{def4}), the close cubes of the global optimum are denoted as $B(x_{1,gbest},\delta_1), B(x_{2,gbest},\delta_2), \cdots, B(x_{D,gbest},\delta_D)$ (please refer to Definition \ref{def6}, suppose the maximum iteration is $T$, the following 1), 2), and 3) hold:

1) According to Definition \ref{def8}, $\mathcal{R}^t \cdot \mathcal{G}^t \cdot f(I_0) = \lbrace \hat{x}_{1,t},\hat{x}_{2,t}, \cdots, \hat{x}_{D,t} \rbrace \subseteq I$ holds. According to Assumption \ref{assumption1} and $D$, $T$ are sufficiently large, there are $B(\hat{x}_{i,t},\delta_{i,t}) \cap B(\hat{x}_{i,j},\delta_{i,j}) \neq \emptyset, 1\leq t,j \leq T, 1 \leq i \leq D$, inferring from Lemma \ref{lem31}:

\begin{equation} \tag{B1} \label{eqB1}
\begin{gathered}
 \forall B(x_{i,gbest},\delta_i) \subseteq I \subseteq \bigcup_{t=1}^T \bigcup_{i=1}^D  B(\hat{x}_{i,t},\delta_{i,t}), 1 \leq t \leq T, 1 \leq i \leq D 
\end{gathered}
\end{equation}

2) According to Definition \ref{def8}, $\mathcal{R}^t \cdot \mathcal{G}^t \cdot f(I_0) = \lbrace \hat{x}_{1,t},\hat{x}_{2,t}, \cdots, \hat{x}_{D,t} \rbrace \subseteq I$ holds. According to Assumption \ref{assumption1}, and $D$, $T$ are sufficiently large, there are $B(\hat{x}_{i,t},\delta_{i,t}) \cap B(\hat{x}_{i,j},\delta_{i,j}) = \emptyset, 1 \leq t \leq T, 1 \leq i \leq D$, inferring from Lemma \ref{lem32}:

\begin{equation} \tag{B2} \label{eqB2}
\begin{gathered}
\forall B(x_{i,gbest},\delta_i) \subseteq I \subseteq \bigcup_{t=1}^T \bigcup_{i=1}^D  B(\hat{x}_{i,t},\delta_{i,t}) \cup E, 1\leq t \leq T, 1 \leq i \leq D\\
\overline{\overline{\textit{E}}}= \aleph_0
\end{gathered}
\end{equation}

3) Otherwise, if some closed cubes are overlapped, and others are not, it needs to consider 1) and 2) comprehensively. In brief, the randomized intervals generated by any optimizer can cover the closed cubes of the global optimum.

\textbf{\textit{Proof}}:
According to \textbf{Definition} \ref{def2}, considering a gradient-based optimizer:
\begin{equation} \tag{B3} \label{eqB3}
\begin{gathered}
    \mathcal{G} \cdot I_0 \subseteq I
\end{gathered}
\end{equation}

According to \textbf{Definition} \ref{def3}, considering a gradient-based optimizer with permutation randomization, we have:
\begin{equation} \tag{B4} \label{eqB4}
\begin{gathered}
      \mathcal{R} \cdot \mathcal{G} \cdot I_0 \subseteq I
\end{gathered}
\end{equation}

According to \textbf{Definition} \ref{def5}, We can derive the following equation:
\begin{equation} \tag{B5} \label{eqB5}
\begin{gathered}
    \mathcal{G}^t \cdot I_0 = \lbrace x_{1,t},x_{2,t}, \cdots, x_{D,t} \rbrace \subseteq I \\
\end{gathered}
\end{equation}

In addition, for any $t \in \mathbb{N}$, we have:
\begin{equation}  \tag{B6} \label{eqB6}
\begin{gathered}
      \mathcal{R}^t \cdot \mathcal{G}^t \cdot I_0 = \lbrace \hat{x}_{1,t},\hat{x}_{2,t},\cdots,x_{D,t} \rbrace \subseteq I
\end{gathered}
\end{equation}

On the one hand, if we consider the overlap through all generated closed cubes via $\mathcal{R}^t \cdot  \mathcal{G}^t \cdot I_0$:
\begin{equation} \tag{B7} \label{eqB7}
\begin{gathered}
      \forall j,k \in \lbrack 1,T \rbrack, j \neq k, B(\hat{x}_{i,j},\delta_{i,j}) \subseteq I,  B(\hat{x}_{i,k},\delta_{i,k} \subseteq \textit{I}, 1 \leq j,k \leq T, 1 \leq i \leq D \\
      B(\hat{x}_{i,j}, \delta_{i,j}) \cap B(\hat{x}_{i,k},\delta_{i,k}) \neq \varnothing\\
\end{gathered}
\end{equation}

Suppose $T$ and $D$ are sufficiently large, inferring from \textbf{Lemma} \ref{lem31}, according to \textbf{Definition} \ref{def6}, if we denote the closed cubes of global optimum as $B(x_{1,gbest},\delta_1), B(x_{2,gbest},\delta_2), \cdots, B(x_{D,gbest}$, we have:
\begin{equation}  \tag{B8} \label{eqB8}
\begin{gathered}
       I \subseteq \bigcup_{t=1}^T \bigcup_{i=1}^D B(\hat{x}_{i,t},\delta_{i,t})  
\end{gathered}
\end{equation}

We can also derive:
\begin{equation} \tag{B9} \label{eqB9}
\begin{gathered}
       \forall  i \in \lbrack 1,D \rbrack, \bigcup_{t=1}^T \bigcup_{i=1}^D B(\hat{x}_{i,t},\delta_{i,t}) \supseteq I \supset B(x_{i,gbest},\delta_i)
\end{gathered}
\end{equation}

On the other hand, if we consider no overlap through all generated intervals via $\mathcal{R}^t \cdot \mathcal{G}^t \cdot I$:
\begin{equation} \tag{B10} \label{eqB10}
\begin{gathered}
      \forall j,k \in \lbrack 1,T \rbrack, j \neq k, B(\hat{x}_{i,j},\delta_{i,j}) \subseteq I, B(\hat{x}_{i,k},\delta_{i,k})  \subseteq I\\
       B(\hat{x}_{i,j},\delta_{i,j}) \cap B(\hat{x}_{i,k},\delta_{i,k}) \neq \varnothing \\
\end{gathered}
\end{equation}

According to \textbf{Lemma} \ref{lem32}, suppose $T$ is sufficiently large, we have:
\begin{equation} \tag{B11} \label{eqB11}
\begin{gathered}
      m^{*}(I \backslash \bigcup_{t=1}^{T} \bigcup_{i=1}^{D}  B(\hat{x}_{i,t},\delta_{i,t})) < \epsilon \\
      E = I \backslash \bigcup_{t=1}^{T} \bigcup_{i=1}^{D} B(\hat{x}_{i,t},\delta_{i,t})\\
       (\bigcup_{t=1}^{T} \bigcup_{i=1}^{D} B(\hat{x}_{i,t},\delta_{i,t})) \bigcup E \supseteq I\\
\end{gathered}
\end{equation}

It also indicates:
\begin{equation} \tag{B12} \label{eqB12}
\begin{gathered}
        \forall i \in \lbrack 1,D \rbrack, (\bigcup_{t=1}^{T} \bigcup_{i=1}^{D} B(\hat{x}_{i,t},\delta_{i,t}) \bigcup E \supseteq I \supseteq B(x_{i,gbest},\delta_i) \\
       \overline{\overline{\textit{E}}}= \aleph_0 \\
\end{gathered}
\end{equation}

Otherwise, it is obvious that $\lbrace B(x_{1, gbest},\delta_1), B(x_{2, gbest},\delta_2), \cdots, B(x_{D, gbest},\delta_D) \rbrace$ can be included in a series of overlapped and non-overlapped closed intervals by comprehensively using Eqs.~\eqref{eqB9} and~\eqref{eqB12}.

\textbf{Lemma} 4.4 Given randomization strategy as an operator denoted as $\mathcal{R}: \mathbb{R}^D \rightarrow \mathbb{R}^D$ (please refer to Definition \ref{def3} in manuscript), we have $\left \| \mathcal{R} \right \| =1$ hold.\\
\textbf{\textit{Proof}}:\\
Considering $\mathcal{R}$ applying on finite-dimensional space:
\begin{equation} \tag{B13} \label{eqB13}
\begin{gathered}
\mathcal{R} \cdot \left[ 
\begin{array}{c}
     x_1\\
     x_2\\
     \vdots\\
     x_D\\
     \end{array}
     \right]
 =  \left[ \begin{array}{c}
     \hat{x}_1\\
     \hat{x}_2 \\
     \vdots\\
     \hat{x}_D\\
     \end{array}
     \right]
\end{gathered}
\end{equation}

Inferring from Eq. \eqref{eqB13}, we have:
\begin{equation} \tag{B14} \label{eqB14}
\begin{gathered}
 \hat{x}_1 = x_i, \hat{x}_2 = x_j, \cdots, \hat{x}_D = x_k, i,j,k \in \lbrack 1,D \rbrack
\end{gathered}
\end{equation}

Eq.~\eqref{eqB2}, we have:
\begin{equation} \tag{B15} \label{eqB15}
\begin{gathered}
 \vert \vert \lbrace x_1, x_2, \cdots, x_D \rbrace \vert \vert = \vert \vert \lbrace \hat{x}_1, \hat{x}_2, \cdots, \hat{x}_D \rbrace \vert \vert
\end{gathered}
\end{equation}

According to the concept of operator norm \citep{Rudin1973functionalanalysis}, we can derive the following:
\begin{equation} \tag{B16} \label{eqB16}
\begin{gathered}
 \vert \vert \mathcal{R} \vert \vert = sup \frac{\mathcal{R} \cdot \vert \vert \lbrace x_1, x_2, \cdots, x_D \rbrace \vert \vert}{\vert \vert \lbrace x_1, x_2, \cdots, x_D \rbrace \vert \vert} = sup \frac{\vert \vert \lbrace \hat{x}_1, \hat{x}_2, \cdots, \hat{x}_D \rbrace \vert \vert}{\vert \vert \lbrace x_1, x_2, \cdots, x_D \rbrace \vert \vert}=1
\end{gathered}
\end{equation}

\textbf{Theorem} 4.5 (Permutation Randomization Preserves the Convergence Rate of the Incorporated Gradient-Based Optimizer) According to Assumption~\ref{assumption1} and Definition~\ref{def8}, the convergence rate of a gradient-based optimizer incorporating randomization should be smaller or equal to the convergence rate of originally incorporated gradient-based optimizer $\mathcal{G}$ without randomization.\\
\textbf{\textit{Proof}}:\\
Inferring from the concept of contraction operator, we have:
\begin{equation} \tag{B17} \label{eqB17}
\begin{gathered}
 \vert \vert \mathcal{G} \cdot (f(X)-f(Y)) \vert \vert \leq c \vert \vert \mathcal{G} \cdot (f(X)-f(Y)) \vert \vert\\
 0<c<1
\end{gathered}
\end{equation}

We can rewrite the left side of Eq. \eqref{eqB16} as:
\begin{equation} \tag{B18} \label{eqB18}
\begin{gathered}
 \vert \vert \mathcal{G} \cdot (f(I_{t+1})-f(I_t)) \vert \vert
\end{gathered}
\end{equation}

Then, we have:
\begin{equation} \tag{B19} \label{eqB19}
\begin{gathered}
 \vert \vert \mathcal{G} \cdot (f(I_{t+1})-f(I_t)) \vert \vert \leq c \cdot \vert \vert (f(I_{t+1})-f(I_t)) \vert \vert
\end{gathered}
\end{equation}

Considering the incorporation of optimizer and randomization as$\mathcal{R} \cdot \mathcal{G} \cdot f(x)$, we have
\begin{equation} \tag{B20} \label{eqB20}
\begin{gathered}
 \vert \vert \mathcal{R} \cdot \mathcal{G} \cdot (f(I_{t+1})-f(I_t)) \vert \vert \leq \vert \vert \mathcal{R} \vert \vert \cdot \vert \vert \mathcal{G} \cdot(f(I_{t+1})-f(I_t)) \vert \vert
\end{gathered}
\end{equation}

Inferring from \textbf{Lemma} \ref{lem33}, it is obvious that we have:
\begin{equation} \tag{B21} \label{eqB21}
\begin{gathered}
\vert \vert  \mathcal{R} \vert \vert \cdot \vert \vert \mathcal{G} \cdot (f(I_{t+1})-f(I_t)) \vert \vert =  \vert \vert \mathcal{G} \cdot(f(I_{t+1})-f(I_t)) \vert \vert \leq c \cdot \vert \vert f(I_{t+1}) -f(I_t) \vert \vert
\end{gathered}
\end{equation}

Eq. \eqref{eqB21} implies permutation randomization $\mathcal{R}$ can maintain the convergence rate of original gradient-based optimizer $\mathcal{G}$.

\textbf{Theorem} 4.6 (Upper Bound on Convergence Rate of Randomized Gradient-based Optimizer Approximates an Closed Cube of Global Optimum) According to Definitions~\ref{def5}, ~\ref{def6}, and Theorem~\ref{thm32}, considering the objective function $\psi:\mathbb{R}^1 \rightarrow \mathbb{R}^1$, if maximum convergence rate is less than $ 2 \cdot \left| \psi(B(x_{gbest},\delta)) \right| $, $\lbrace x, x_{gbest} \in I \subseteq \mathbb{R}^D, \psi(x) \leq \psi(B(x_{gbest},\delta)) \rbrace  = \bigcup_{k=1}^T \bigcup_{m=1}^T \bigcap_{t=m}^T \lbrace x \in I \subseteq \mathbb{R}^D: max \lbrack \mathcal{R}^t \cdot \mathcal{G}^t \cdot \psi(x) \rbrack + \frac{1}{k}\rbrace$ holds.

\textbf{\textit{Proof}}:\\
Let $E_{t,k}=\lbrack x \in I: max_{x \in I} \mathcal{R}^t \cdot \mathcal{G}^t \cdot \psi(x) \leq \psi(B(x_{gbest},\delta)) + \frac{1}{k}\rbrace$.

On the one hand, we need to prove $\lbrace x, x_{gbest} \in \mathbb{R}^1: \psi(x) \leq \psi(B(x_{gbest},\delta)) \rbrace \subseteq \bigcup_{k=1}^T \bigcup_{m=1}^T \bigcap_{t=m}^T \lbrace x \in \mathbb{R}^1: max_{x \in I} \mathcal{R}^t \cdot \mathcal{G}^t \cdot \psi(x) \leq \psi(B(x_{gbest},\delta)) + \frac{1}{k}\rbrace$.

Suppose $x_{gbest} \in \mathbb{R}^1$ and $x_0 \in \lbrace x, x_{gbest} \in \mathbb{R}^1: \psi(x) \leq \psi(B(x_{gbest},\delta)) \rbrace$, according to \textit{Cauchy Theorem}, we have $lim_{t \rightarrow \infty} \psi_t (x_0) = \psi(x_0) \leq \psi(B(x_{gbest},\delta))$ and derive:
\begin{equation} \tag{B22} \label{eqB22}
\begin{gathered}
\forall k_0 \in \mathbb{N}, \exists t_0 \geq t_0, \psi_t (x_0) \leq \psi(B(x_{gbest},\delta)) + \frac{1}{k_0}
\end{gathered}
\end{equation}

Eq.~\eqref{eqB10} indicates that $x_0 \in E_{t,k_0} (t \geq t_0)$. It also demonstrates $x_0$ belongs to the lower limit of $\lbrace E_{t,k_0} \rbrace$.

According to the Definition of the lower limit, we have:
\begin{equation} \tag{B23} \label{eqB23}
\begin{gathered}
\forall x_0 \in I, x_0 \in \bigcup_{m=1}^T \bigcap_{t=m}^T E_{t,k_0}
\end{gathered}
\end{equation}

Thus, we proved $\lbrace x, x_{gbest} \in I \subseteq \mathbb{R}^D: f(x) \leq f(B(x_{gbest},\delta)) \rbrace \subseteq \bigcup_{k=1}^T \bigcup_{m=1}^T \bigcap_{t=m}^T \lbrace x \in I \subseteq \mathbb{R}^D: max_{x \in I} \mathcal{R}^t \cdot \mathcal{G}^t \cdot f(x) \leq f(B(x_{gbest},\delta)) + \frac{1}{k}\rbrace$.

On the other hand, we need to prove $\lbrace x, x_{gbest} \in I \subseteq \mathbb{R}^D: f(x) \leq f(B(x_{gbest},\delta)) \rbrace \supseteq \bigcup_{k=1}^T \bigcup_{m=1}^T \bigcap_{t=m}^T \lbrace x \in I \subseteq \mathbb{R}^D: max_{x \in I} \mathcal{R}^t \cdot \mathcal{G}^t \cdot f(x) \leq f(B(x_{gbest},\delta)) + \frac{1}{k}\rbrace$.

Suppose $x_0 \in \bigcup_{k=1}^T \bigcup_{m=1}^T \bigcap_{t=m}^T \lbrace x \in I \subseteq \mathbb{R}^D: max_{x \in \mathbb{R}^1} \mathcal{R}^t \cdot \mathcal{G}^t \cdot f(x) \leq f(B(x_{gbest},\delta)) + \frac{1}{k}\rbrace$, obviously, we have $x_0 \in \bigcup_{m=1}^T \bigcap_{t=m}^T \lbrace x \in I \subseteq \mathbb{R}^D: max_{x \in \mathbb{R}^1} \mathcal{R}^t \cdot \mathcal{G}^t \cdot f(x) \leq \psi(B(x_{gbest},\delta)) + \frac{1}{k}\rbrace$.

Importantly, it implies that $x_0 \in \bigcup_{m=1}^T \bigcap_{t=m}^T R_{t,K_0}$ which is the lower limit of $\lbrace E_{t,k_0} \rbrace$.

Thus, $\exists t_0, x_0 \in E_{t,k_0}, t \geq t_0$, we can derive:
\begin{equation} \tag{B24} \label{eqB24}
\begin{gathered}
\psi_t (x_0) \leq \psi(B(x_{gbest},\delta)) + \frac{1}{k_0} ( t \geq t_0 )
\end{gathered}
\end{equation}

Let $t$ and $k_0$ are sufficiently large, we have $f(x_0) \leq f(B(x_{gbest},\delta))$.

So far, we proved $\lbrace x, x_{gbest} \in \mathbb{R}^1: \psi(x) \leq \psi(B(x_{gbest},\delta)) \rbrace \supseteq \bigcup_{k=1}^T \bigcup_{m=1}^T \bigcap_{t=m}^T \lbrace x \in \mathbb{R}^1: max_{x \in I} \mathcal{R}^t \cdot \mathcal{G}^t \cdot \psi(x) \leq \psi(B(x_{gbest},\delta)) + \frac{1}{k}\rbrace$.

Moreover, according to \textbf{Theorem}~\ref{thm32} and Eq. \eqref{eqB12}, we have:
\begin{equation} \tag{B25} \label{eqB25}
\begin{gathered}
\vert max_{x \in I} \lbrace \mathcal{R}^t \cdot \mathcal{G}^t \cdot \psi(x) \rbrace - max_{x \in \mathbb{R}^1} \lbrace \mathcal{R}^{t+1} \cdot \mathcal{G}^{t+1} \cdot \psi(x) \rbrace \vert \leq 2 \cdot \vert f(x_{gbest}, \delta) \vert 
\end{gathered}
\end{equation}

Therefore, we can finally conclude that the upper bound of convergence rate of a gradient-based optimizer in each dimensionality incorporating permutation randomization as:
\begin{equation} \tag{B26} \label{eqB26}
\begin{gathered}
\frac{1}{\sqrt{T}} \leq 2 \cdot \vert f(x_{i, gbest}, \delta) \vert, 1 \leq i \leq D\\
\end{gathered}
\end{equation}

\newpage
\section{Appendix C: Additional Experiments}
\begin{figure}[htbp]
  \centering
  \includegraphics[width=0.86\textwidth]{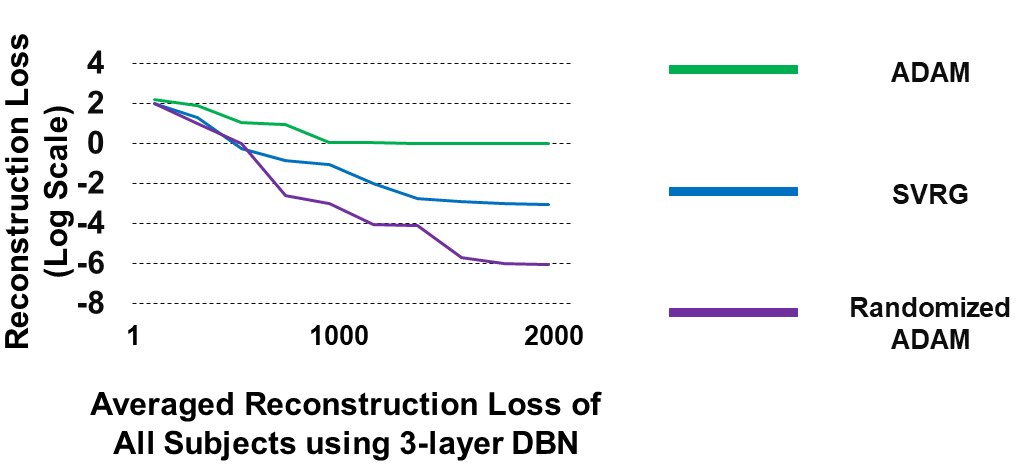}
  \caption{An illustration of reconstruction loss comparisons of randomized ADAM and other peer optimizers on optimizing a 3-layer DBN.}
  \label{fig:fig6}
\end{figure}

Figure~\ref{fig:fig6} presents a comparison between randomized ADAM and two other leading optimizers, ADAM and SVRG, in optimizing a three-layer deep belief network (DBN). Notably, randomized ADAM achieves the highest reconstruction accuracy among the methods compared.

\begin{figure}[htbp]
  \centering
  \includegraphics[width=0.86\textwidth]{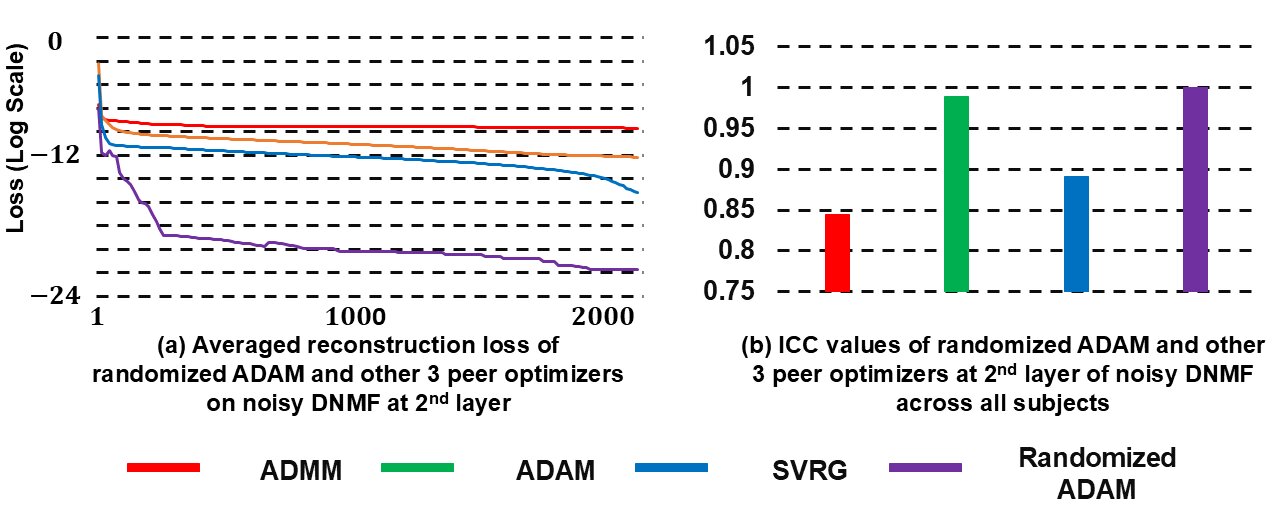}
  \caption{An illustration of reconstruction loss comparisons of randomized ADAM and other peer optimizers.}
  \label{fig:fig7}
\end{figure}

Additionally, Figure~\ref{fig:fig7} shows the average reconstruction loss for randomized ADAM and two other peer optimizers over 2000 iterations in the optimization of noisy Deep Nonlinear Matrix Factorization (DNMF). Specifically, this model introduces noise into the gradient $g_t$ at every 10 iterations to enhance the complexity of the optimization process.

\begin{figure}[htbp]
  \centering
  \includegraphics[width=0.96\textwidth]{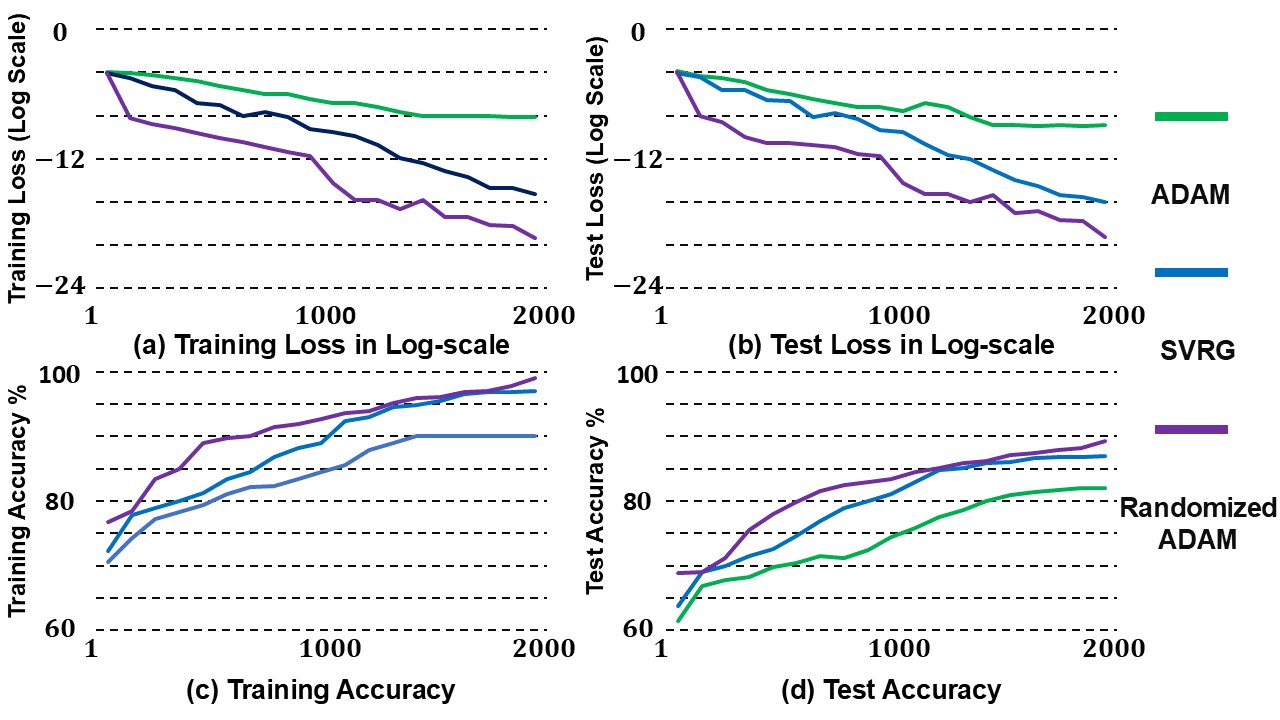}
  \caption{An illustration of reconstruction loss comparisons of randomized ADAM and other peer optimizers on solving the logistic regression problem.}
  \label{fig:fig8}
\end{figure}

Finally, we provide the pseudo-code of randomized ADAM below:
\begin{table}[H]
  \caption{The Pseudo Code of Randomized ADAM}
  \centering
  \begin{tabular}{l}
    \toprule
   \textbf{Algorithm 1 randomized ADAM:} Stochastic ADAM\\
    \midrule
    \textbf{Input:} $\alpha$ denotes step size; $\beta_1$ and $\beta_2$ are exponential decay;\\
    $\theta_0$ is initial parameter vector; $m_0$ is initialized 1st moment vector;\\
    $v_0$ is initialized 2nd moment vector;\\
    $t$ is current iteration;$\epsilon$ is initialized as a threshold; $T$ is the maximum iteration.\\
    \textbf{while} $t<T$\\
    \,\, $g_t \leftarrow \nabla_x f(x_{t})$\\
    \,\, $M_t \leftarrow \beta_1 M_{t-1} + (1-\beta_1)g_t$\\
    \,\, $V_t \leftarrow \beta_2 V_{t-1} + (1-\beta_1)g_t$\\
    \,\, $\hat{M}_t \leftarrow \frac{M_t}{1-\beta_1^t}$\\
    \,\, $\hat{V}_t \leftarrow \frac{V_t}{1-\beta_2^t}$\\
    \,\, $x_t \leftarrow x_{t-1} - \alpha \cdot \frac{\hat{M}_t}{\sqrt{\hat{V}_t}+\epsilon}$\\
     \,\, \textbf{if} $\left \| g_t-g_{t-1}\right \| < \epsilon$\\
     \,\,\,\, $\hat{x}_t \leftarrow \mathcal{R}(x_t)$\\
     \,\,\,\, $x_{t} \leftarrow \hat{x}_t$\\
     \,\, \textbf{End if}\\
     \,\, $\textit{t} \leftarrow \textit{t}+1$\\
     \textbf{End while}\\
    \bottomrule
  \end{tabular}
  \label{table:table3}
\end{table}

\end{document}